%% file: main.tex
\documentclass[10pt,twocolumn,letterpaper]{article}

\usepackage[pagenumbers]{cvpr} 
\usepackage{graphicx}
\usepackage{amsmath}
\usepackage{amssymb}
\usepackage{booktabs}
\usepackage{bigstrut}
\usepackage{array}
\usepackage{colortbl}
\usepackage[T1]{fontenc}
\usepackage{wrapfig,lipsum}
\usepackage{rotating}
\usepackage{multirow}
\usepackage{placeins}
\usepackage{gensymb}
\usepackage{xcolor}
\usepackage{soul}
\usepackage{makecell}
\makeatletter  
\@namedef{ver@everyshi.sty}{}
\makeatother
\usepackage{tikz}
\usepackage{yfonts}
\usepackage{import}
\usepackage{pifont} 
\usepackage{appendix}
\usepackage[accsupp]{axessibility}  
\usepackage{hhline}
\usepackage{stfloats}
\usepackage{setspace}
\usepackage{wasysym}
\usepackage{atbegshi,picture}   
\usepackage{mathtools}
\usepackage[font={footnotesize}]{caption}

\definecolor{cvprblue}{rgb}{0.21,0.49,0.74}
\usepackage[pagebackref,breaklinks,colorlinks,allcolors=cvprblue]{hyperref}
\newcommand{\boldparagraphstart}[1]{\vspace{1pt}\noindent \textbf{#1}}
\newcommand{\xmark}{\text{\ding{55}}}  
\definecolor{darkgreen}{RGB}{0,127,0}
\definecolor{darkred}{RGB}{200,0,0}
\def\greencheckmark{\textcolor{darkgreen}{\checkmark}}
\def\redxmark{\textcolor{darkred}{\xmark}}

\def\authorspace{\quad\quad}

\newcommand*{\affmark}[1][*]{\textsuperscript{#1}}

\newenvironment{myitem}{\begin{list}{$\bullet$}
{\setlength{\itemsep}{-0pt}
\setlength{\topsep}{0pt}
\setlength{\labelwidth}{5pt}
\setlength{\leftmargin}{10pt}
\setlength{\parsep}{-0pt}
\setlength{\itemsep}{0pt}
\setlength{\partopsep}{0pt}}}%
{\end{list}}

\def\confName{CVPR}
\def\confYear{2026}
\begin{document}

\title{Fast-FoundationStereo: Real-Time
Zero-Shot Stereo Matching}

\author{Bowen Wen\affmark[] \authorspace Shaurya Dewan\affmark[] \authorspace Stan Birchfield\affmark[]\\[8pt] 
\affmark[]NVIDIA
}

\twocolumn[{
\renewcommand\twocolumn[1][]{#1}%
\maketitle

\begin{center}
    \centering
     \vspace{-25pt}
     \includegraphics[width=0.99\textwidth]{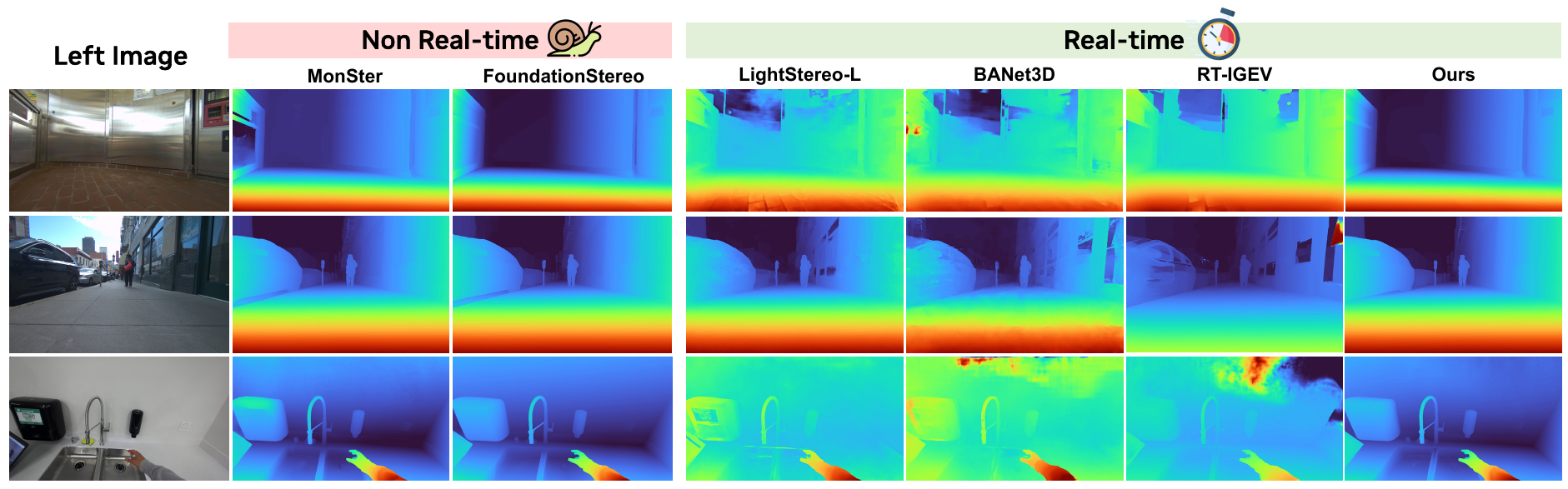}
      \vspace{-0.15in}
    \captionof{figure}{Our \textit{Fast-FoundationStereo} achieves comparable results to MonSter~\cite{cheng2025monster} and FoundationStereo~\cite{wen2025foundationstereo} while running nearly 10 times faster.  Shown are disparity maps obtained by zero-shot inference on in-the-wild images.  (Note that our results occasionally exceed those of \cite{cheng2025monster}, \eg, the shiny door in the top row, and the paper towel bin in the bottom row.)} \label{fig:intro}
\end{center}%
}]

\begin{abstract}
Stereo foundation models achieve strong zero-shot
generalization but remain computationally prohibitive for
real-time applications.  Efficient stereo architectures, on the other hand, sacrifice
robustness for speed and require costly per-domain fine-tuning.
To bridge this gap, we present Fast-FoundationStereo, a family of architectures that achieve, for the first time, strong zero-shot generalization at real-time frame rate. We employ a divide-and-conquer acceleration strategy with three components: (1) knowledge distillation to compress the hybrid backbone into a single efficient student; (2) blockwise neural architecture search for automatically discovering optimal cost filtering designs under latency budgets, reducing search complexity exponentially; and (3) structured pruning for eliminating redundancy in the iterative refinement module. Furthermore, we introduce an automatic pseudo-labeling pipeline used to curate 1.4M in-the-wild stereo pairs to supplement synthetic training data and facilitate knowledge distillation. The resulting model can run over 10× faster than FoundationStereo while closely matching its zero-shot accuracy, thus establishing a new state-of-the-art among real-time methods. Project page: \url{https://nvlabs.github.io/Fast-FoundationStereo/}
\end{abstract}
\vspace{-20pt}

\section{Introduction}
The field of stereo matching has advanced significantly since its inception exactly 50 years ago~\cite{marrpoggio1976}.

Modern algorithms, driven by an abundance of high-quality training datasets and innovations in deep neural network architectures, now yield impressive results, often approaching saturation on the most demanding benchmarks.
Such accuracy is critical for applications requiring precise 3D reconstruction, such as robotics~\cite{lee2025delta} and augmented reality~\cite{kong2025aria}. 

\begin{figure}
    \centering
     \vspace{-10pt}
     \includegraphics[width=0.9\columnwidth]{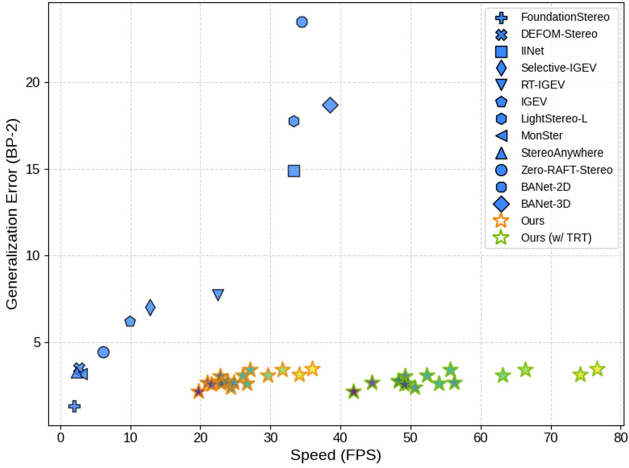}
      \vspace{-0.15in}
    \caption{Zero-shot generalization accuracy (on Middlebury-Q dataset) of various stereo methods versus speed, measured on the same hardware NVIDIA 3090 GPU. Our model family achieves real-time performance with only slight decrease in accuracy compared with the best slow method. Green outlined stars are further accelerated by TensorRT.} \label{fig:introplot}
    \vspace{-20pt}
\end{figure}

This remarkable progress, however, has split the field into two distinct research paths~\cite{Tosi_survey_IJCV2025}. On the one hand, the rise of foundation models in computer vision has pushed stereo research toward strong zero-shot generalization~\cite{wen2025foundationstereo,wang2025zerostereo,cheng2025monster,bartolomei2025stereo}. 
Such leading zero-shot networks leverage rich priors from computationally intensive foundation models such as DepthAnythingV2~\cite{yang2024depthanythingv2} or DINO models~\cite{oquab2023dinov2,simeoni2025dinov3}; and they employ computationally intensive architectures, such as the Disparity Transformer~\cite{wen2025foundationstereo}, to perform self-attention for long-range context. Such limitations have, to date, hindered their deployment in any latency-bound system.

On the other hand, the non-negotiable constraints of practical applications demand computationally efficient performance. 
Architectures designed for such real-time inference~\cite{li2024iinet,guo2025lightstereo,xu2025banet,igevpp} achieve high frame rates by relying on lightweight backbones, 2D convolutional layers, and local iterative refinement modules. Such methods struggle to generalize due to their reliance upon per-domain fine-tuning.  The difficulty of obtaining the required dense, high-quality ground-truth depth at scale has prevented such efficient methods from being used as an off-the-shelf solution for embodied agents operating in  in-the-wild environments.

To address this critical gap, we propose Fast-FoundationStereo (Fig.~\ref{fig:intro}), a novel stereo matching approach for both strong zero-shot generalization and real-time inference. Unlike existing real-time methods which sacrifice the rich architectural capacity and typically designed and trained from scratch, our work builds upon the powerful yet computationally intensive FoundationStereo~\cite{wen2025foundationstereo}.  Addressing its three main components (feature extraction, cost filtering, and disparity refinement), our divide-and-conquer acceleration strategy takes into account the unique properties of each. First, knowledge distillation is leveraged to compress the computationally expensive hybrid feature backbone into a single, efficient student backbone that retains the rich monocular and stereo priors. Second, the intensive cost filtering network is divided into blocks, numerous candidate blocks are trained via distillation, and combinatorial search automatically discovers a family of effective architectures under varying latency budgets. Third, structured pruning is applied to the refinement module, guided by a recurrent dependency graph to identify and remove redundancy, followed by retraining to recover performance. Finally, training is supplemented with a large-scale (1.4M pairs) dataset of in-the-wild stereo images, curated via an automatic pseudo-labeling pipeline.

Our contributions can be summarized as follows:
\begin{myitem}
    \item We present Fast-FoundationStereo, a novel stereo matching architecture that achieves both strong zero-shot generalization and real-time inference, with varying accuracy-speed trade-off (Fig.~\ref{fig:introplot}).  Our method significantly outperforms other real-time models by a large margin across multiple public datasets, and even outperforms several recent strongly generalizable models.
    \item We present several novelties to address the computational bottleneck of common components adopted in modern stereo matching models, while inheriting the strengths from the teacher model. Our divide-and-conquer strategy includes:  (1) distillation from hybrid monocular and stereo priors, (2) cost filtering via efficient blockwise architecture search, and (3) iterative refinement via structured pruning.
    \item To harness the large diversity, internet-scale abundance  and unique realism from in-the-wild stereo images, we propose an automatic pseudo-labeling pipeline to supplement synthetic training data for knowledge distillation.
\end{myitem}

\section{Related Work}

\boldparagraphstart{Generalizable Stereo Matching.} Recent progress in generalizable stereo matching has centered on leveraging Vision Foundation Models (VFMs) and monocular priors to achieve strong zero-shot performance. FoundationStereo~\cite{wen2025foundationstereo} establishes a strong baseline by adapting DepthAnythingV2 with side-tuning, while StereoAnywhere~\cite{bartolomei2025stereo} demonstrates robustness where stereo or mono cues fail independently, and MonSter~\cite{cheng2025monster} marries monocular depth with stereo matching to unleash complementary strengths. 
ZeroStereo~\cite{wang2025zerostereo} synthesizes additional training data based on monocular depth estimation and diffusion models. DEFOM-Stereo~\cite{jiang2025defom} builds upon depth foundation models, All-in-One~\cite{zhou2025all} systematically transfers VFMs into stereo frameworks, and recent work diving into the fusion of monocular priors~\cite{yao2025diving} analyzes effective integration strategies. Beyond direct adaptation, domain generalization has been pursued through domain-invariant representations~\cite{zhang2020domain}, learning from foundation models for domain generalized stereo matching~\cite{zhang2025learning}, and information-theoretic approaches that avoid shortcut learning~\cite{chuah2022itsa}. Additional architectural innovations include hierarchical visual transformations~\cite{chang2023domain}, masked representation learning for domain generalized stereo matching~\cite{rao2023masked}, and harnessing broad-spectrum task-oriented features~\cite{liu2022graftnet}. Despite impressive zero-shot generalization, their computational overhead remains prohibitive for real-time applications.

\boldparagraphstart{Efficiency-Oriented Stereo Matching.} 
Efficiency-oriented stereo matching architectures have traditionally pursued real-time performance through three primary strategies: compact cost volume representations, lightweight processing modules, and streamlined network designs. The first strategy reduces memory footprint via low-resolution feature pyramids~\cite{khamis2018stereonet}, 2D cost signatures~\cite{yee2020fast}, attention-based disparity selection~\cite{xu2022attention}, or learned parameterized functions that replace explicit volumes entirely~\cite{zeng2023parameterized}. The second strategy accelerates cost aggregation by pruning search spaces in coarse-to-fine cascades~\cite{gu2020cascade}, operating in efficient bilateral grid spaces~\cite{xu2021bilateral}, or employing 3D separable convolutions~\cite{rahim2021separable} to avoid expensive 3D kernels. The third strategy designs mobile-specific architectures~\cite{shamsafar2022mobilestereonet}, tile-based iterative refinement~\cite{tankovich2021hitnet}, or binary operations~\cite{cai2022pbcstereo} from the ground up, while more sophisticated approaches employ neural architecture search to automatically discover efficient networks~\cite{cheng2020hierarchical,wang2022easnet}. While these methods achieve impressive frame rates, they fundamentally sacrifice the rich architectural capacity. In addition, the models are usually designed and trained from scratch, ignoring  recent powerful foundation models. Consequently, they remain tethered to per-domain fine-tuning on target distributions, making them unsuitable solutions for in-the-wild environments.

\begin{figure*}[t]
    \centering
    {\includegraphics[width=0.95\textwidth]{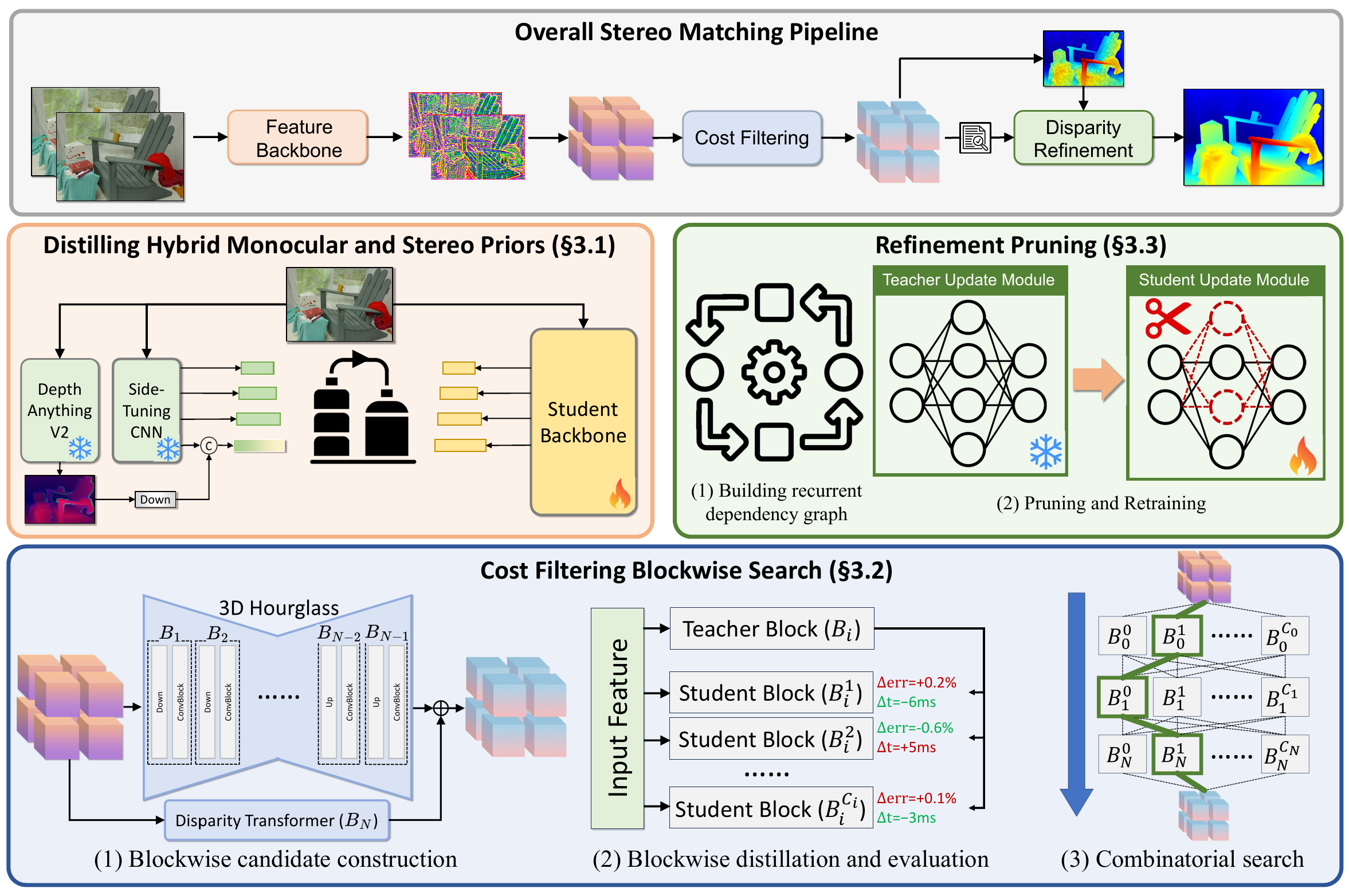}} 
    \vspace{-10pt}
    \caption{Overview of our framework. \textbf{Top:} Foundational stereo matching networks (\eg,~\cite{wen2025foundationstereo}) consist of three key steps: feature extraction, cost filtering, and disparity refinement. Each step is accelerated by a divide-and-conquer strategy. \textbf{Middle-Left:} Hybrid monocular and stereo priors from the teacher foundation model are distilled into a single backbone student model. \textbf{Middle-right:} Refinement network is pruned by first constructing a dependency graph that models the recurrent nature of the GRU module, followed by structured pruning and retraining to recover the accuracy. \textbf{Bottom:} Cost filtering network is divided into separate local blocks; block candidates are trained to match the teacher block's output, taking as input the local feature from the previous block; and combinatorial search finds the best performing block combination for a given runtime constraint.}\label{fig:pipeline}
    \vspace{-10pt}
\end{figure*}

\boldparagraphstart{Vision Foundation Model Acceleration.} The significant computational overhead of Vision Foundation Models (VFMs) has spurred a large body of research focused on their acceleration for practical deployment. A recent active area has been the optimization of SAM~\cite{kirillov2023segment} and VGGT~\cite{wang2025vggt}, with several distinct approaches. Many works propose efficient architectures, introducing entirely new lightweight models or modifying existing ones for speed~\cite{fu2024lite,xiong2024efficientsam,zhang2024efficientvit}. Another common strategy is quantization, which reduces numerical precision to speed up inference, as demonstrated by PTQ4SAM~\cite{lv2024ptq4sam} and Quantized-VGGT~\cite{feng2025quantized}. Methods like SlimSAM~\cite{chen2024slimsam} employ structured pruning, followed by  distillation to create highly compact models. Knowledge distillation is also often used independently to transfer knowledge from a large teacher model to a smaller student~\cite{zhou2025edgesam,zhang2023faster}. Finally, some methods leverage domain-specific knowledge to accelerate computationally expensive components, such as Fast-VGGT~\cite{shen2025fastvggt} which uses token merging. In comparison, accelerating large foundation models for stereo matching has been largely under-explored, leaving a substantial research gap.

\section{Approach}

Our approach (Fig.~\ref{fig:pipeline}) is based on FoundationStereo~\cite{wen2025foundationstereo}, which consists of three key steps: feature extraction, cost filtering, and disparity refinement. Each of these steps is accelerated by a divide-and-conquer strategy, as detailed in the following subsections.  
We also describe our automatic data curation pipeline.

\subsection{Distilling Hybrid Monocular and Stereo Priors}

 Given a pair of  left and right images $I_l, I_r\in \mathbb{R}^{H\times W \times 3}$, the feature backbone extracts multi-level pyramid features $ f_l^{(i)}, f_r^{(i)} \in \mathbb{R}^{C_i \times \frac{H}{i} \times \frac{W}{i}} $, $i \in \{4, 8, 16, 32\}$ for the subsequent cost volume construction and aggregation. To compute such features, FoundationStereo~\cite{wen2025foundationstereo} combines DepthAnything V2~\cite{yang2024depthanythingv2} 
 with 
 a side-tuning CNN.  The former provides rich monocular priors learned from large-scale internet data, and the latter adapts the monocular features for the binocular stereo setup. Although such hybrid monocular and stereo feature extraction is powerful, it remains a significant computational bottleneck. 
 
We leverage knowledge distillation to replace the dual module in FoundationStereo's backbone with a single student module.
This approach was chosen because it is agnostic to architecture and allows to build upon the well-established feature backbones studied on ImageNet~\cite{rw2019timm,deng2009imagenet}.
As an alternative, we also considered model pruning, but it has two drawbacks:  (1) it would require us to keep the dual module, which is constrained by the computational bottleneck of its underlying ViT~\cite{dosovitskiy2020image}; and (2) any deterioration in accuracy would be difficult to recover without retraining on internet-scale imagery. 
 
\begin{figure}[t]
    \centering
    {\includegraphics[width=0.45\textwidth]{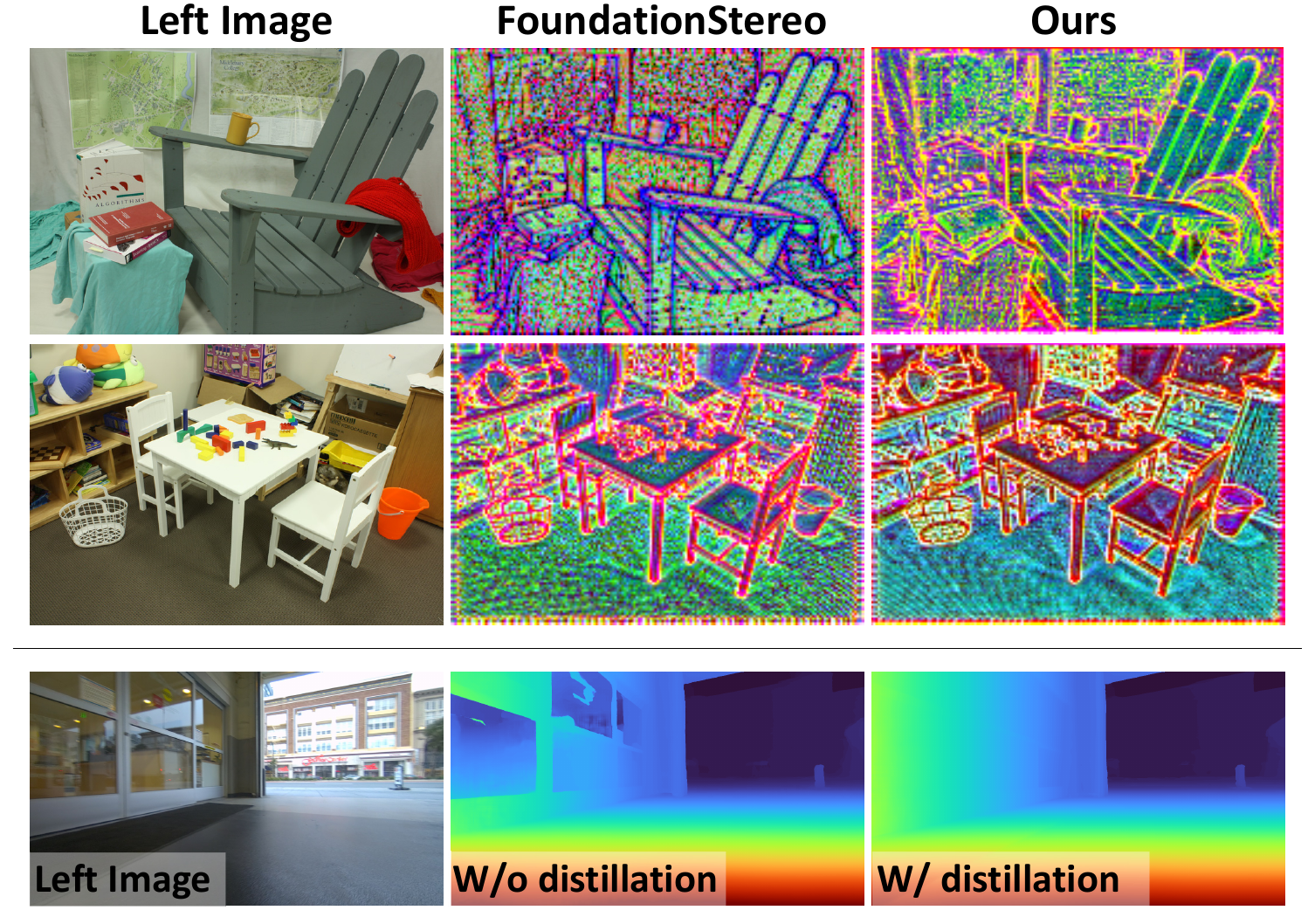}} 
    \vspace{-10pt}
    \caption{\textbf{Top:} Distilling the hybrid monocular and stereo priors from FoundationStereo~\cite{wen2025foundationstereo} into a unified single backbone captures similar high-frequency edges and relative depth---while significantly reducing computational cost. \textbf{Bottom:} Distillation enhances robustness to translucency, which is challenging to traditional stereo matching.} \label{fig:backbone_feat}
    \vspace{-20pt}
\end{figure}

During distillation, DepthAnything V2 and side-tuning CNN modules from FoundationStereo are frozen and used to predict a multi-level feature pyramid $\bar{f}^{(i)}$, which the student model is trained to match via MSE loss. In the case of channel dimension mismatch, a linear projection layer is added. Even though the feature extractors take only a single image as input, we include both stereo images in each training batch to retain the statistical similarity.  
To provide a family of stereo models with different speed-accuracy trade-off, we train multiple variants of feature extractors~\cite{maaz2022edgenext, sandler2018mobilenetv2}. 
Fig.~\ref{fig:backbone_feat}
 visualizes examples of distilled features, showing that they  capture similar high-frequency edges and relative depth.
 
\subsection{Cost Filtering Blockwise Search}\label{sec:cost_filtering}

Given the unary features extracted in the previous step, the cost volume $\mathbf{V_C} \in \mathbb{R}^{C\times \frac{D}{4} \times \frac{H}{4} \times \frac{W}{4}}$ is constructed by combining the group-wise correlation and concatenation volumes (where $D$ is the maximum disparity). To effectively scale the learning process with the abundant training data, FoundationStereo~\cite{wen2025foundationstereo} uses a dual branch architecture to perform cost filtering. Specifically, a 3D hourglass architecture consisting of Axial-Planar Convolution (APC) layers effectively processes $\mathbf{V_C}$ by enlarging the kernel size over the disparity dimension without significantly increasing memory consumption. Meanwhile, a Disparity Transformer branch tokenizes $\mathbf{V_C}$ and performs multi-head self-attention to further enhance the long-range context reasoning within the 4D cost volume. 

Direct pruning of the cost filtering modules  yields severe performance degradation for only marginal speedup, since the channel dimension in $\mathbf{V_C}$ is already small (mostly under 100). We avoided direct knowledge distillation, because it requires manually designing the cost-filtering module alternatives, which remain less explored than feature backbones. Instead, we leverage Neural Architecture Search (NAS)~\cite{elsken2019neural} to automatically discover non-intuitive  designs. In the following, we describe our efficient blockwise search strategy  (Fig.~\ref{fig:pipeline} bottom).

\boldparagraphstart{Blockwise Candidate Construction.} The cost filtering module is divided into a series of operation blocks: $\Phi_t(\mathbf{V_C})=B_{N} \circ \cdots  \circ B_2 \circ B_1(\mathbf{V_C})$, where $N$ represents the total number of blocks. Within the 3D hourglass module, blocks are divided at the transition of the channel dimension, which typically corresponds to the spatial dimension change of the feature volume. 
We define five types of layers: (1) 3D conv layer with varying channel dimensions; (2) 3D deconv layer that doubles the spatial dimensions of the cost volume, (3) APC layer~\cite{wen2025foundationstereo} that performs separate spatial and disparity convolution with different respective kernel sizes; (4) residually connected 3D conv layers, similar to ResNet~\cite{he2016deep}; and (5) feature guided volume excitation~\cite{bangunharcana2021correlate}. 

Meanwhile, the entire Disparity Transformer module is regarded as a single block consisting of a number of repeated multi-head self-attention transformer layers. 
We reuse the disparity attention layers as in \cite{wen2025foundationstereo} while varying the feed-forward layer dimensions, number of heads, and number of layers.  

In both cases, the number of layers in each block and the intermediate channel dimension can vary, as long as (1) the entire block's running time $t_B^s$ is faster than its teacher counterpart $t_B^t$ and (2) the input and output channel dimension remains the same as the original block. Details of the search space can be found in the appendix.

\boldparagraphstart{Blockwise Distillation and Evaluation.}
After blockwise candidate construction,  we obtain $C=C_1 \cdot C_2 \cdots C_N$ total number of possible cost filtering module candidates, where $C_i$ denotes the number of candidates in block $B_i$. In practice, when $N=8$ and $C_i=200$, $C$ is $200^8\approx 10^{18}$. As a result, standard evolutionary search based NAS methods~\cite{desell2017large,real2019regularized} are not tractable, due to the extremely large computational cost. Moreover, training from scratch in the whole search space does not fully leverage the strengths of the teacher model. 

To overcome these limitations, we train each block $B_i$ independently. Specifically, $B_i$ is treated as a standalone network and trained to mimic the teacher counterpart's output: $\left\| B_i(f_{i-1})-\bar{B}_i(f_{i-1}) \right\|^2_2$, given the feature output $f_{i-1}$ from the previous teacher block. For the final block that predicts the initial disparity, smooth $L_1$ loss is computed against the ground truth. The teacher model is frozen throughout the distillation process. 
After distillation, a candidate block $B_i^c$ is evaluated by replacing its counterpart at block level $i$ in the teacher model and inferring the complete model end-to-end on a separate validation dataset. Both the relative error metric change $\Delta m_i^c$ and running time change $\Delta t_i^c$ that result by introducing $B_i^c$ are measured.

Compared with standard NAS, our blockwise distillation reduces training complexity from $O(n^N)$ to $O(n)$~\cite{moons2021distilling,li2020block}, where $n$ is the number of per-layer candidates. Furthermore, since $B_i$ is small, the block distillation can be performed efficiently in terms of both speed and memory, allowing easy parallelization. 

\boldparagraphstart{Combinatorial Search.} The student cost filtering module is found by solving for the optimal combination of candidate blocks, which can be formulated as:
\begin{align}\label{eq:ilp}
\min_{\cal{E}} \sum_{i=1}^{N} (\Delta \mathbf{m}_i)^\top \mathbf{e}_i, \quad \text{s.t.} \quad \sum_{i=1}^{N}  (\Delta \mathbf{t}_i)^\top \mathbf{e}_i \leq \Delta \tau,
\end{align}
where $\Delta \mathbf{m}_i$ and $\Delta \mathbf{t}_i$ denote the vector of error metric and running time  changes, respectively, for all candidates at block $B_i$; $\mathbf{e}_i \in \cal{E}$ denotes the one-hot vector representing the selection operation of a candidate block at $B_i$; and $\Delta \tau$ denotes the runtime budget relative to the teacher model for the entire cost filtering module. Optimization is performed by Integer Linear Programming (ILP)~\cite{mitchell2011pulp,molchanov2022lana}, using different values for $\tau$ to obtain a family of cost filtering student models with different speed-accuracy trade-off.

\subsection{Refinement Pruning}

\begin{figure}[t]
    \centering
    {\includegraphics[width=0.45\textwidth]{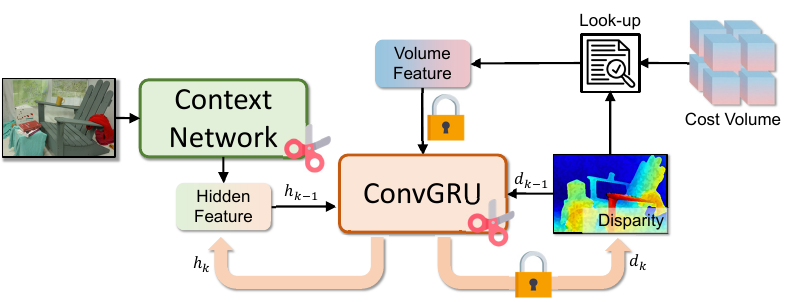}} 
    \vspace{-10pt}
    \caption{Recurrent dependency graph of the refinement module. \includegraphics[height=1em]{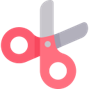} denotes where pruning is performed. \includegraphics[height=1em]{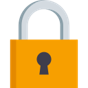} denotes where channel dimension remains fixed during the pruning process.}\label{fig:prune}
    \vspace{-20pt}
\end{figure}

Given the initial disparity map $d_0$ (predicted by the filtered cost volume) and the hidden feature (initialized from the context network), the ConvGRU module progressively refines the disparity map. Fig.~\ref{fig:prune} shows the dependency graph and data flow. At each iteration, ConvGRU module consumes the disparity $d_{k-1}$, $h_{k-1}$ and predicts their updated values $d_k, h_k$, resulting in recurrent dependencies. 
This significant redundancy in refinement module  (as shown in Sec.~\ref{sec:framework}), motivates the use of structured pruning~\cite{he2023structured,molchanov2016pruning,fang2023depgraph,muralidharan2024compact}, a simple yet effective technique and can benefit from GPU hardware acceleration techniques such as TensorRT. 

\boldparagraphstart{Building Recurrent Dependency Graph.} The first step in structured pruning is to identify the inter-dependencies between layers, since depth or channel pruning at one layer changes the intermediate feature dimensions fed to adjacent layers.  In addition to the normal adjacent layer dependencies which can be automatically constructed by tracing the computation flow~\cite{fang2023depgraph}, we introduce three more pruning constraints given the unique properties of refinement module in stereo matching: (1) within the ConvGRU module, the final layers that predict the disparity map and convex upsampling mask retain fixed output channel dimensions; (2) within the ConvGRU module, the input channel of the layer that consumes $h_{k-1}$, and the output channel of the layer that outputs $h_k$, are inter-dependent and thus jointly pruned; and (3) the motion encoder that consumes the indexed volume feature retains a fixed input channel dimension. 

\boldparagraphstart{Pruning and Retraining.} To identify which layers or channels to remove, we evaluate their importance using first-order Taylor expansion~\cite{molchanov2019importance}. Specifically, inputs are feed forward to the complete teacher model~\cite{wen2025foundationstereo} end-to-end with multiple refinement iterations, and accumulate gradients for the refinement module. The importance of each parameter in the refinement module is ranked globally, and the least important $\alpha$ parameters are pruned, where $\alpha \in (0,1)$ is the pruning ratio. We also explored isomorphic pruning strategy~\cite{fang2024isomorphic} but observed slightly degraded performance. After pruning, we retrain the refinement module end-to-end (while freezing the rest of the teacher model) to recover the performance, using the loss: 
\begin{align}\label{eq:prune_retrain}
    \mathcal{L}=\sum_{k=1}^{K}\gamma^{K-k}\left\| d_k-\overline{d} \right\|_1 + \lambda\sum_{i=1}^{L}\left\| x_i-\overline{x}_i  \right\|_2^2
\end{align}
where $x_i$ and $\overline{x}_i$ are the per-layer latent features (student and teacher, respectively) from each of the $L$ layers; $\overline{d}$ is the ground truth disparity; $k$ is the iteration number; $\gamma=0.9$ exponentially increases weights to supervise the iteratively refined disparity; and $\lambda=0.1$ weighs the distillation objective. The initial disparity supervision is excluded since it is not affected by the refinement module.

\begin{figure}[t]
    \centering
    {\includegraphics[width=0.45\textwidth]{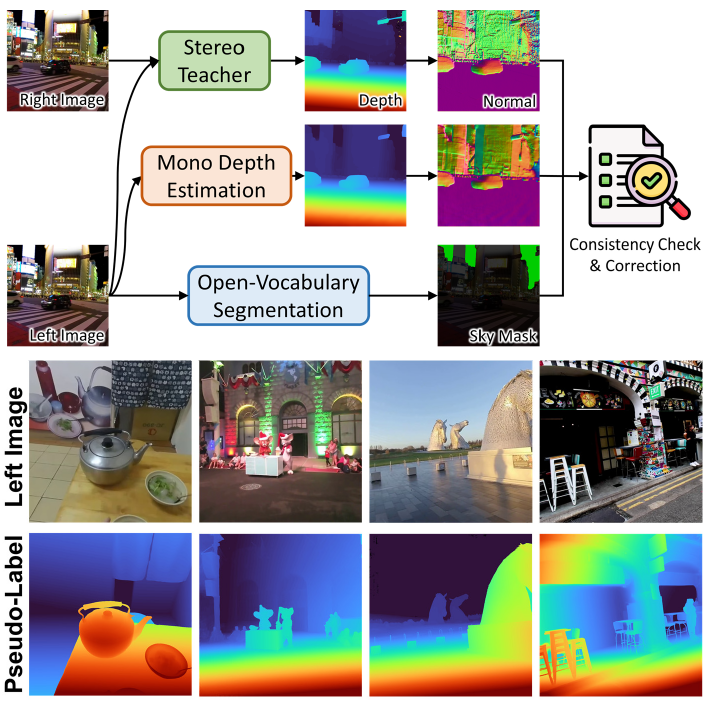}} 
    \vspace{-10pt}
    \caption{\textbf{Top:} Pseudo-labeling pipeline on in-the-wild internet stereo data. \textbf{Bottom:} Visualization of our generated pseudo-labels.}\label{fig:stereo4d}
    \vspace{-20pt}
\end{figure}

\subsection{Pseudo-Labeling on In-the-Wild Data}\label{sec:psuedo}
Real-world data offers greater diversity and realism than synthetic data. However, obtaining real stereo images with ground-truth metric depth annotation is notoriously difficult. To address this challenge, we propose an automatic data curation pipeline to generate pseudo-labels on internet-scale stereo images. As shown in Fig.~\ref{fig:stereo4d}, given a rectified stereo pair from Stereo4D~\cite{stereo4d}, the teacher model~\cite{wen2025foundationstereo}  produces a disparity map for the left image. To identify the imperfect predictions which can mislead the subsequent training process for the student model, we also feed left image to a monocular depth estimator~\cite{piccinelli2025unidepthv2} to obtain a corresponding depth map. Both the disparity map and monocular depth are further converted into normal maps via 3D unprojection and Sobel operator using the same set of camera parameters provided by \cite{stereo4d}. To assess local geometric consistency, we compute the per-pixel cosine similarity between the two normal maps, which is thresholded to produce a consistency mask. Stereo samples with insufficient agreement are discarded. 
Due to the uniqueness of sky regions (which have infinite depth and are underrepresented in common synthetic datasets used for training), the similarity computation excludes the sky regions, which are detected by  open-vocabulary segmentation models~\cite{ravi2024sam,xu2023open}.

The remaining stereo disparity maps become the final pseudo-labels, where the sky regions are set to zero disparity. The consistency mask can be optionally used to determine the supervision pixels. We subsample the videos temporally by a stride of 10, yielding 1.4M suitable stereo pairs in total. In contrast to directly comparing in the depth or disparity space, our proposed normal consistency check is more robust to extremely diverse depth ranges or noisy predictions on in-the-wild images.  This automatically pseudo-labeled data is included in our final training of student models. Such output-space distillation complements the feature-based distillation performed in previous steps.

\section{Experiments}

\input{table/zero_shot}

\begin{figure*}[t]
    \centering
    {\includegraphics[width=0.99\textwidth]{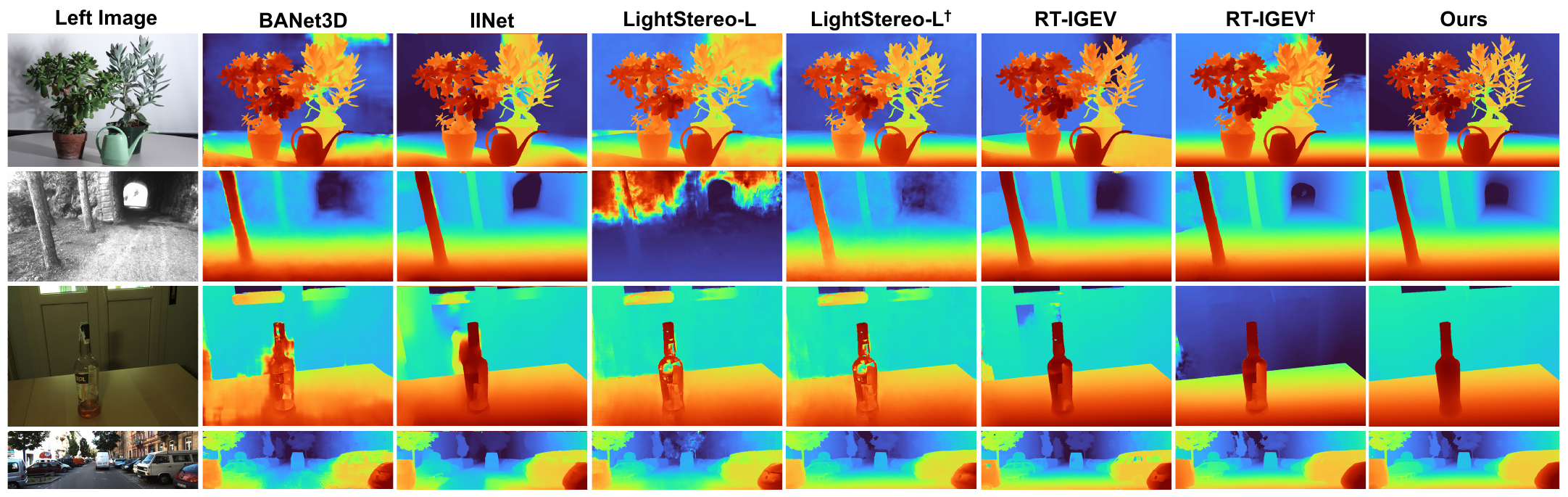}} 
    \vspace{-10pt}
    \caption{Qualitative results of real-time methods on Middlebury, ETH3D, Booster and KITTI-2015 datasets (top to bottom). Results are obtained by zero-shot inference without training on any split of the target datasets. $^\dagger$Indicates methods trained on the exact same datasets as ours, including our proposed pseudo-labels. More results on in-the-wild images can be found in the appendix.}\label{fig:zero_shot}
    \vspace{-10pt}
\end{figure*}

\input{table/booster}

\subsection{Implementation Details}\label{sec:impl}

Our Fast-FoundationStereo was trained on the same mixed datasets as FoundationStereo~\cite{wen2025foundationstereo}, as well as the pseudo-labeled real data (Sec.~\ref{sec:psuedo}). For deployment, our framework provides the flexibility to assemble different candidates from each step to compose the final stereo matching model (examples are shown in Fig.~\ref{fig:introplot}). The final model is then trained end-to-end. Once trained, the fixed set of weights for each candidate model are used to perform zero-shot inference on unseen data. Unless otherwise mentioned, we use 8 refinement iterations and 192 as the maximum disparity for constructing the cost volume. For evaluation, the disparity range is not constrained.

\subsection{Benchmark Datasets and Metric}\label{sec:datasets}

\boldparagraphstart{Datasets.} Four common  public datasets were used for evaluation: 
Middlebury~\cite{middlebury} consists of indoor stereo image pairs  with high-quality ground-truth disparity captured via structured light.
ETH3D~\cite{eth3d} provides grayscale stereo image pairs covering both indoor and outdoor scenarios. and
KITTI 2012~\cite{geiger2012we} and KITTI 2015~\cite{menze2015object} feature real-world driving scenes, where sparse ground-truth disparity maps are derived from LIDAR sensors. Booster~\cite{ramirez2023booster} features a large variety of translucent and specular scenes and is used to evaluate the robustness to non-Lambertian surfaces.

\boldparagraphstart{Metrics.} ``BP-X'' computes the percentage of pixels where the disparity error is larger than X pixels. ``D1'', commonly used on KITTI~\cite{geiger2012we,menze2015object},  computes the percentage of pixels whose disparity error is larger than 3 pixels and 5\% of the ground-truth disparity. Results are evaluated on non-occluded regions.

\subsection{Zero-Shot Generalization Comparison}

\boldparagraphstart{Quantitative Comparison.} 
Comparison of zero-shot generalization on public datasets is shown in Table~\ref{tab:zero_shot}. These datasets are unseen to all the evaluated methods. For cost-filtering based methods that support dynamic maximum disparity configuration, 416 is used on Middlebury-H for the best performance; otherwise their default setting is used for other lower resolution datasets.
Existing real-time methods are usually not targeted for zero-shot generalization, and are thus mainly trained on SceneFlow~\cite{sceneflow2016}. For those competitive ones with publicly released training code~\cite{guo2025lightstereo,igevpp}, we additionally train them on the exact same datasets as ours (including our proposed pseudo-labels). The inference runtimes for all methods are profiled over Middlebury-Q (similar to typical resolution for real-time robotic applications) on the same hardware with NVIDIA 3090 GPU. 

As can be observed, our Fast-FoundationStereo outperforms other real-time models by a significant margin across the board, even when they are trained on the exact same datasets, including our proposed pseudo-labels. Moreover, our model achieves comparable or even better results than most of the  computationally expensive models, including Zero-RAFT-Stereo~\cite{wang2025zerostereo} which leverages additional synthesized training data via multiple large foundation models. Compared to FoundationStereo~\cite{wen2025foundationstereo}, our method runs more than 10 times faster with only a modest increase in error. 

\boldparagraphstart{Robustness to Non-Lambertian Surfaces.} Table~\ref{tab:booster} shows zero-shot generalization results on Booster-Q dataset~\cite{ramirez2023booster}. Numbers are from the StereoAnywhere paper~\cite{bartolomei2025stereo}; FoundationStereo~\cite{wen2025foundationstereo} and the most competitive real-time model RT-IGEV~\cite{igevpp} are also included for comparison.

\boldparagraphstart{Qualitative Comparison.} Visualizations of zero-shot inference are demonstrated in Figs.~\ref{fig:intro} and~\ref{fig:zero_shot}. The stereo images represent diverse challenges including textureless regions, transparency, specular highlights, complex illuminations, varying depth ranges, viewing perspectives and both indoor / outdoor scenarios. Despite these challenges, our model significantly outperforms other real-time models. It even achieves comparable or sometimes more favorable results than computationally expensive generalizable models.

\subsection{Framework Analysis}\label{sec:framework}

\boldparagraphstart{Effects of Backbone Distillation.} Table~\ref{tab:backbone} shows an ablation study on no distillation (feature backbone weights pretrained only on ImageNet~\cite{deng2009imagenet}) and different distillation losses. By distilling from hybrid monocular and stereo priors from the teacher model, the feature backbone generally enhances zero-shot generalization. The effectiveness is also demonstrated in Fig.~\ref{fig:backbone_feat}, where the translucent glass door challenges the traditional stereo matching process without distillation.

\input{table/backbone_distill}

\begin{figure}[h]
    \centering
    {\includegraphics[width=0.47\textwidth]{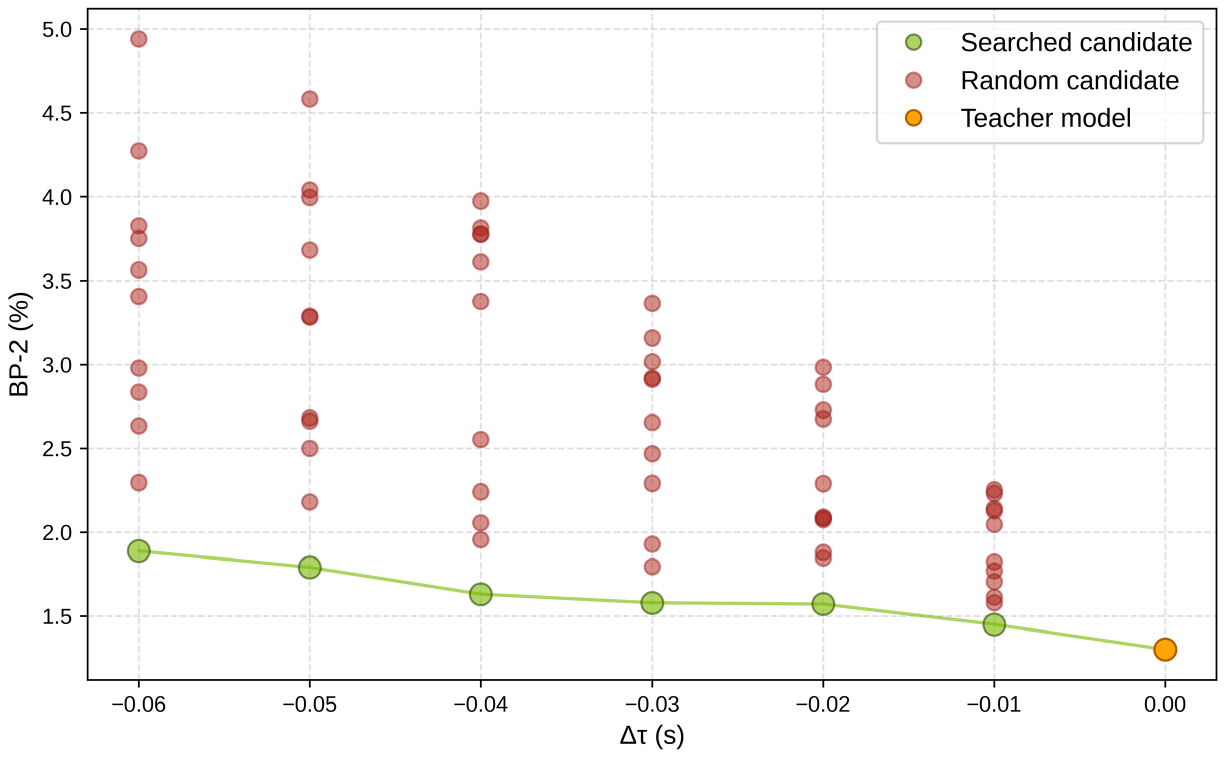}} 
    \vspace{-10pt}
    \caption{Effects of blockwise architecture search for cost filtering module under varying latency budget $\Delta \tau$, evaluated on Middlebury-Q.} \label{fig:blockwise_search}
\end{figure}

\boldparagraphstart{Effects of Cost Filtering Blockwise Search.} Our blockwise search strategy significantly reduces training complexity from $O(n^N)$ to $O(n)$. However, it leverages a surrogate objective, Eq.~\eqref{eq:ilp}, which accumulates the impacts of perturbing each local block. This is a proxy to the actual performance of a candidate model, which would otherwise require training the full assembled cost filtering module with the remaining parts of the network end-to-end for evaluation. In order to verify if such proxy is effective, we compare our searched cost filtering candidate, based on Eq.~\eqref{eq:ilp}, against randomly assembled candidates under the same latency constraint $\Delta \tau$. All cost filtering module candidates are trained end-to-end (with the remaining parts from the teacher model), followed by zero-shot evaluation. For each $\Delta \tau$, 10 random candidate models are sampled (note that training each of them end-to-end is expensive). As shown in Fig.~\ref{fig:blockwise_search}: (1) as the latency constraint relaxes ($\Delta \tau$ increases), our architecture search can successfully find better performing candidate models; (2) under varying $\Delta\tau$, the searched candidate consistently outperforms randomly assembled candidates; and (3) as $\Delta\tau$ decreases, some randomly assembled candidates yield substantial performance degradation, highlighting the importance of network design under tight latency constraint.

\begin{figure}[h]
    \centering
    {\includegraphics[width=0.43\textwidth]{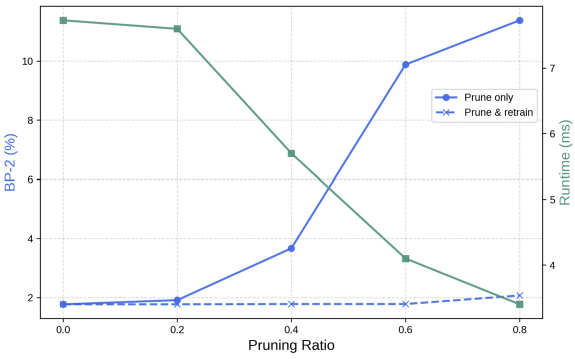}} 
    \vspace{-10pt}
    \caption{Effects of pruning ratio for accuracy and speed.} \label{fig:pruning_line}
    \vspace{-15pt}
\end{figure}

\boldparagraphstart{Effects of Pruning Ratio.} Fig.~\ref{fig:pruning_line} demonstrates how the pruning ratio affects the prediction accuracy on Middlebury-Q dataset and runtime under one refinement iteration. While aggressive pruning dramatically degrades the prediction accuracy, it can be effectively recovered through retraining with Eq.~\eqref{eq:prune_retrain}, implying large redundancy in the original refinement module.

\boldparagraphstart{Effects of Pseudo-Labeling.} Table~\ref{tab:psuedo_label} ablates on training with pseudo-labeled data and their zero-shot generalization results on public datasets under commonly used metrics. We also ablate on a few competitive real-time methods with this data. Pseudo-labeling enhances generalization performance consistently for all methods. The elevation is even more significant for those methods~\cite{igevpp,guo2025lightstereo} that were previously trained only on SceneFlow.
\input{table/psuedo_label}

\begin{figure}[h]
    \centering
    \vspace{-10pt}
    {\includegraphics[width=0.45\textwidth]{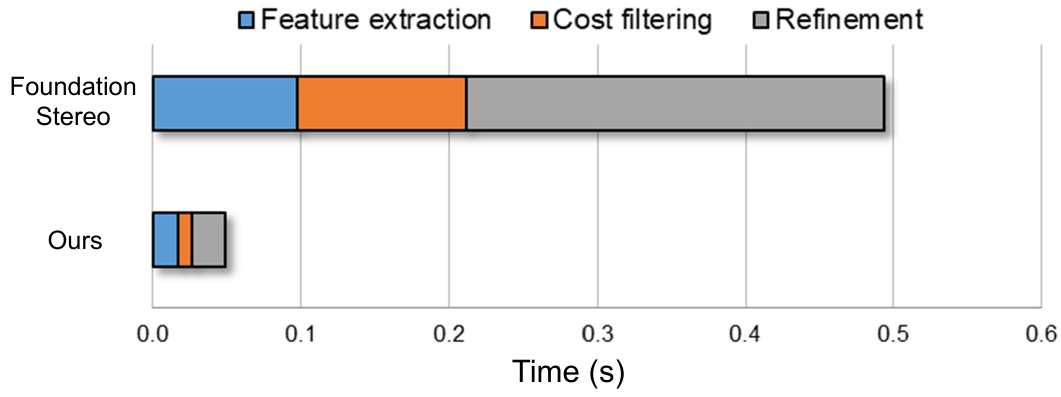}} 
    \vspace{-10pt}
    \caption{Runtime decomposition.} \label{fig:runtime}
    \vspace{-20pt}
\end{figure}

\boldparagraphstart{Runtime Analysis.} Fig.~\ref{fig:runtime} shows the detailed runtime decomposition between FoundationStereo~\cite{wen2025foundationstereo} and our slowest model from Fig.~\ref{fig:introplot}. Results are profiled on the same hardware (NVIDIA 3090 GPU). Each of the three essential steps are accelerated by a large margin, leading to a total runtime performance boost of more than 10$\times$.

\section{Conclusion}
Fast-FoundationStereo bridges the gap between zero-shot generalization and real-time performance. Through a principled divide-and-conquer acceleration strategy, we demonstrate that the computational bottlenecks of foundation stereo models can be systematically addressed without sacrificing robustness. 
Our extensive evaluations confirm that Fast-FoundationStereo not only establishes a new state-of-the-art among real-time methods by a substantial margin, but it also competes favorably with computationally intensive generalizable models. 
For future work, exploring quantization techniques offers an orthogonal avenue to further enhance inference speed, potentially enabling deployment on even more resource-constrained edge devices.

{
    \small
    \bibliographystyle{ieeenat_fullname}
    \bibliography{ref}
}

\input{suppl}

\end{document}

%% file: table/zero_shot.tex
\begin{table*}[t]
\centering
\def\mywidth{0.98\textwidth} 
\definecolor{green}{RGB}{0,200,0}
\resizebox{\mywidth}{!}{

\begin{tabular}{lrrrrrrrrrrrrrrrrrc}
\toprule
\multirow{2}[2]{*}{Method} & \multicolumn{3}{c}{Middlebury-H} & \multicolumn{3}{c}{Middlebury-Q} & \multicolumn{3}{c}{ETH3D} & \multicolumn{4}{c}{KITTI 2012} & \multicolumn{4}{c}{KITTI 2015} & \multirow{2}[2]{*}{Runtime (ms)} \\
      & \multicolumn{1}{c}{BP-1} & \multicolumn{1}{c}{BP-2} & \multicolumn{1}{c}{BP-3} & \multicolumn{1}{c}{BP-1} & \multicolumn{1}{c}{BP-2} & \multicolumn{1}{c}{BP-3} & \multicolumn{1}{c}{BP-1} & \multicolumn{1}{c}{BP-2} & \multicolumn{1}{c}{BP-3} & \multicolumn{1}{c}{BP-1} & \multicolumn{1}{c}{BP-2} & \multicolumn{1}{c}{BP-3} & \multicolumn{1}{c}{D1} & \multicolumn{1}{c}{BP-1} & \multicolumn{1}{c}{BP-2} & \multicolumn{1}{c}{BP-3} & \multicolumn{1}{c}{D1} &  \\
      \cmidrule(lr){1-1} \cmidrule(lr){2-4} \cmidrule(lr){5-7} \cmidrule(lr){8-10} \cmidrule(lr){11-14} \cmidrule(lr){15-18} \cmidrule(lr){19-19}
\rowcolor[rgb]{ .906,  .902,  .902} StereoAnywhere~\cite{bartolomei2025stereo} & 9.67  & 4.75  & 2.45  & 8.00  & 3.25  & 2.10  & 1.43  & 0.61  & 0.41  & 11.66 & 4.67  & 3.52  & 2.81  & 21.81 & 6.72  & 3.79  & 3.52  & 427 \\
DEFOM-Stereo~\cite{jiang2025defom} & 8.84  & 3.76  & 2.46  & 7.51  & 3.50  & 2.22  & 2.16  & 1.03  & 0.78  & 13.10 & 5.32  & 3.39  & 3.12  & 23.92 & 8.12  & 4.76  & 4.58  & 371 \\
\rowcolor[rgb]{ .906,  .902,  .902} MonSter~\cite{cheng2025monster} & 9.33  & 4.24  & 2.69  & 7.08  & 3.19  & 1.94  & 0.99  & 0.46  & 0.28  & 9.58  & 4.39  & 2.99  & 2.84  & 20.61 & 6.44  & 3.59  & 3.41  & 336 \\
Zero-RAFT-Stereo~\cite{wang2025zerostereo} & 8.48  & 4.68  & 3.32  & 8.15  & 4.42  & 3.26  & 2.14  & 1.17  & 0.85  & 9.15  & 4.17  & 2.93  & 2.76  & 21.13 & 7.43  & 4.67  & 4.48  & 164 \\
\rowcolor[rgb]{ .906,  .902,  .902} FoundationStereo~\cite{wen2025foundationstereo} & \textbf{2.49}  & \textbf{1.10}  & \textbf{0.88}  & \textbf{2.64}  & \textbf{1.30}  & \textbf{0.96}  & \textbf{0.50}  & \textbf{0.30}  & \textbf{0.24}  & \textbf{8.16}  & \textbf{3.50}  & \textbf{2.47}  & \textbf{2.30}  & \textbf{18.65} & \textbf{5.20}  & \textbf{2.95}  & \textbf{2.80}  & 496 \\
      \cmidrule(lr){1-1} \cmidrule(lr){2-4} \cmidrule(lr){5-7} \cmidrule(lr){8-10} \cmidrule(lr){11-14} \cmidrule(lr){15-18} \cmidrule(lr){19-19}
IINet$^*$~\cite{li2024iinet} & 25.88 & 16.69 & 13.03 & 24.90 & 14.90 & 10.42 & 21.21 & 12.55 & 9.19  & 33.12 & 15.71 & 9.72  & 9.30  & 36.22 & 14.16 & 7.86  & 7.58  & \,\,\,30 \\
\rowcolor[rgb]{ .906,  .902,  .902} LightStereo-L$^*$~\cite{guo2025lightstereo} & 37.49 & 23.76 & 18.48 & 30.08 & 17.75 & 13.11 & 45.46 & 37.21 & 34.15 & 42.42 & 22.39 & 14.49 & 13.98 & 40.56 & 19.10 & 12.35 & 12.08 & \,\,\,30 \\
LightStereo-L~\cite{guo2025lightstereo} & 22.64 & 12.55 & 9.07  & 16.34 & 7.70  & 4.99  & 16.34 & 7.70  & 4.99  & 17.59 & 6.71  & 3.97  & 3.73  & 27.66 & 9.07  & 4.75  & 4.51  & \,\,\,30 \\
\rowcolor[rgb]{ .906,  .902,  .902} RT-IGEV$^*$~\cite{igevpp} & 16.95 & 11.52 & 9.40  & 14.02 & 7.71  & 5.52  & 5.66  & 2.81  & 2.26  & 16.70 & 7.28  & 4.85  & 4.54  & 25.89 & 9.89  & 6.19  & 6.00  & \,\,\,45 \\
RT-IGEV~\cite{igevpp} & 12.75& 7.82 & 5.73  & 11.28 & 5.59 & 3.77  & 5.05  & 2.78  & 1.63  & 11.38 & 5.05  & 3.44  & 3.25  & 22.70 & 7.32  & 4.24  & 4.00  & \,\,\,45 \\
\rowcolor[rgb]{ .906,  .902,  .902} BANet-2D$^*$~\cite{xu2025banet} & 43.78 & 28.45 & 22.28 & 37.33 & 23.51 & 18.62 & 44.89 & 37.87 & 35.81 & 42.92 & 22.45 & 14.48 & 13.88 & 42.92 & 22.45 & 14.48 & 13.88 & \,\,\,29 \\
BANet-3D$^*$~\cite{xu2025banet} & 44.90 & 30.10 & 24.17 & 32.02 & 18.69 & 13.70 & 29.27 & 20.99 & 18.55 & 45.43 & 25.20 & 17.07 & 16.59 & 50.26 & 26.38 & 17.13 & 16.87 & \,\,\,26 \\
\rowcolor[rgb]{ .906,  .902,  .902} Ours  &  \textbf{4.80}  &  \textbf{2.20}     &  \textbf{1.60}     &  \textbf{4.51}     &   \textbf{2.12}    &  \textbf{1.57}     &  \textbf{1.22}     &   \textbf{0.62}    &  \textbf{0.50}     &  \textbf{8.52}     &  \textbf{3.61}     &  \textbf{2.50}     &  \textbf{2.35}     & \textbf{19.62}      & \textbf{5.78}      &  \textbf{3.43}     &  \textbf{3.25}     & \,\,\,\,\,\,\,\,\,\,\,\,\,\,\,49 (21)  \\
\bottomrule
\end{tabular}%

}
\vspace{-8pt}
\caption{Zero-shot generalization on public datasets. Methods are grouped based on their feasibility for real-time application. $^*$Denotes methods trained only on SceneFlow~\cite{sceneflow2016}; others are trained on large-scale combined datasets, or they leverage foundation models pretrained on large-scale datasets.  Bold indicates the best method within each group; note that ours is also second-best in each column.  The number in parentheses is the runtime using TensorRT.}
\label{tab:zero_shot}
\vspace{-10pt}
\end{table*}

%% file: table/booster.tex
\begin{table}[t]
\centering
\def\mywidth{0.48\textwidth} 
\definecolor{green}{RGB}{0,200,0}
\resizebox{\mywidth}{!}{

\begin{tabular}{lrrrrrc}
\toprule
Methods & BP-2  & BP-4  & BP-6  & BP-8  & EPE (px) & Real-time \\
\midrule
RAFT-Stereo~\cite{lipson2021raft} & 17.84 & 13.06 & 10.76 & 9.24  & 3.59  & \redxmark \\
PSMNet~\cite{chang2018pyramid} & 34.47 & 24.83 & 20.46 & 17.77 & 7.26  & \redxmark \\
GMStereo~\cite{gmstereo} & 32.44 & 22.52 & 17.96 & 15.02 & 5.29  & \redxmark \\
PCVNet~\cite{zeng2023parameterized} & 22.63 & 16.51 & 13.81 & 12.08 & 4.70  & \redxmark \\
DLNR~\cite{zhao2023high} & 18.56 & 14.55 & 12.61 & 11.22 & 3.97  & \redxmark \\
Selective-IGEV~\cite{selective_stereo} & 18.52 & 14.24 & 12.14 & 10.77 & 4.38  & \redxmark \\
IGEV~\cite{igev} & 16.90 & 13.23 & 11.40 & 10.20 & 3.94  & \redxmark \\
NMRF~\cite{nmrf} & 27.08 & 19.06 & 15.43 & 13.21 & 5.02  & \redxmark \\
StereoAnywhere~\cite{bartolomei2025stereo} & 9.01  & 5.40  & 4.12  & 3.34  & 1.21  & \redxmark \\
FoundationStereo~\cite{wen2025foundationstereo} & \textbf{5.18} & \textbf{4.07} & \textbf{2.91} & \textbf{2.59} & \textbf{1.13} & \redxmark \\
\midrule
RT-IGEV~\cite{igevpp} & 23.09 & 16.86 & 14.10 & 12.47 & 5.03  & \greencheckmark \\
RT-IGEV$^{\dagger}$~\cite{igevpp} & 18.19 & 13.39 & 11.37 & 10.16 & 4.20  & \greencheckmark \\
Ours  & \textbf{6.61} & \textbf{4.62} & \textbf{3.91} & \textbf{3.49} & \textbf{1.54} & \greencheckmark \\
\bottomrule
\end{tabular}%

}
\vspace{-8pt}
\caption{Zero-shot generalization on non-Lambertian  surfaces, evaluated on the Booster-Q dataset~\cite{ramirez2023booster}. $^\dagger$Denotes training on the exact same datasets as ours (including our proposed pseudo-labels).}\label{tab:booster}
\vspace{-15pt}
\end{table}

%% file: table/backbone_distill.tex
\begin{table}[h]
\centering
\vspace{-5pt}
\def\mywidth{0.45\textwidth} 
\definecolor{green}{RGB}{0,200,0}
\resizebox{\mywidth}{!}{

\begin{tabular}{lrrrr}
\toprule
\multirow{2}[2]{*}{Variants} & Midd.-H &  ETH3D & KITTI-12  & KITTI-15  \\
      & BP-2  & BP-1  & D1    & D1 \\
\midrule
No Distillation & 2.87  & 2.11  & 2.67  & 4.32 \\
Cosine Similarity & 2.29  & \textbf{1.19}  & 2.39  & 3.31 \\
MSE (Ours)   & \textbf{2.20}  & 1.22  & \textbf{2.35}  & \textbf{3.25} \\
\bottomrule
\end{tabular}%

}
\vspace{-8pt}
\caption{Ablations on feature backbone distillation strategies.}
\label{tab:backbone}
\vspace{-10pt}
\end{table}

%% file: table/psuedo_label.tex




\begin{table}[h]
\centering
\vspace{-10pt}
\def\mywidth{0.48\textwidth} 
\definecolor{green}{RGB}{0,200,0}
\resizebox{\mywidth}{!}{
\begin{tabular}{lrrrr}
\toprule
\multirow{2}[2]{*}{Method} & Midd.-H & ETH3D & KITTI-12 & KITTI-15 \\
      & BP-2  & BP-1  & D1    & D1 \\
\midrule
RT-IGEV~\cite{igevpp} & 11.52 \,\,\,(8.69) & 5.66 \,\,\,(5.12) & 4.54  (3.55) & 6.00  (4.40) \\
LightStereo-L~\cite{guo2025lightstereo} & 23.76 (18.41) & 45.46 (21.12) & 13.98 (5.27) & 12.08 (7.63) \\
Ours  & 2.53 \,\,\,(2.20) & 1.31 \,\,\,(1.22) & 2.44 (2.35) & 3.48 (3.25) \\
\bottomrule
\end{tabular}%

}
\vspace{-8pt}
\caption{Results on in-the-wild data without (and with) pseudo-labeling.}\label{tab:psuedo_label}
\vspace{-10pt}
\end{table}

%% file: suppl.tex
\clearpage
\setcounter{page}{1}
\maketitlesupplementary

\section{Real-time Demo}
Please watch our supplemental video which demonstrates real-time demos running on GPU 3090, using the model evaluated in Table~\ref{tab:zero_shot} (main paper).

\section{More Details on Cost Filtering}
We now elaborate the search space for blockwise neural architecture search for cost filtering module. Within 3D hourglass, the following layer options are considered:
\begin{myitem}
    \item \textbf{3D conv layer.} Output channel dimensions are set to $0.5\times, 1\times$ or $2\times$ of the input channel dimension. Kernel size is set to 3. Stride is set to 2 unless downsampling is performed where it is set to 1. Batch normalization and activation operations are optionally included.
    \item \textbf{3D deconv layer.} This layer is chosen when upsampling the spatial dimensions of the cost volume. The parameters are set following the original counterparts in FoundationStereo~\cite{wen2025foundationstereo}.
    \item \textbf{APC layer.}  Output channel dimensions are set to $0.5\times$ or $1\times$ of the input channel dimension. Kernel size of axial-convolution is chosen from ${3,9,17}$. Kernel size of planar-convolution is set to 3. Stride is set to 1.
    \item \textbf{Residually connected 3D conv layers.} We follow the Basic Block structure in ResNet~\cite{he2016deep}, where two convolutional layers of kernel size 3 are residually connected. Output channel dimensions are set to $0.5\times$ or $1\times$ of the input channel dimension. 
    \item \textbf{Feature guided volume excitation.} The multi-level unary features for the left image $f_l^{(i)} \in \mathbb{R}^{C_i \times \frac{H}{i} \times \frac{W}{i}} , i \in \{4, 8, 16, 32\}$ are provided as guidance to excite the relevant geometric features in the cost volume~\cite{bangunharcana2021correlate}.
\end{myitem}
Within the Disparity Transformer, the self-attention transformer encoder layer repeats from 1 to 6 times. The hidden dimension of feedforward layer is chosen between $2\times$ or $4\times$ of the input feature dimension. The number of heads is set to 2 or 4.

In each block the number of layers are chosen to be no more than the number of layers in original teacher's counterpart. In total we divided cost filtering module into $N=8$ blocks. By assembling the different blocks' candidates, we obtain $5.5 \times 10^{24}$ number of cost filtering module designs, where one of them is the original cost filtering module from teacher model. By considering only the design choices faster than teacher, we obtain $5.8\times 10^{19}$ possible combinations. Nevertheless, with our introduced blockwise distillation (Sec.~\ref{sec:cost_filtering} in main paper), only 2584 blocks need to be trained, which can be performed efficiently in terms of both time and memory, allowing ease of parallelization. The whole blockwise distillation process takes 14 days in total distributed on 128 NVIDIA A100 GPUs. After distillation, the most promising block combinations are identified via ILP which can be efficiently solved under a second. For the model evaluated in Table~\ref{tab:zero_shot} (main paper), the latency budget was set to $\Delta\tau=-0.04s$.

\begin{figure}[h]
    \centering
    {\includegraphics[width=0.48\textwidth]{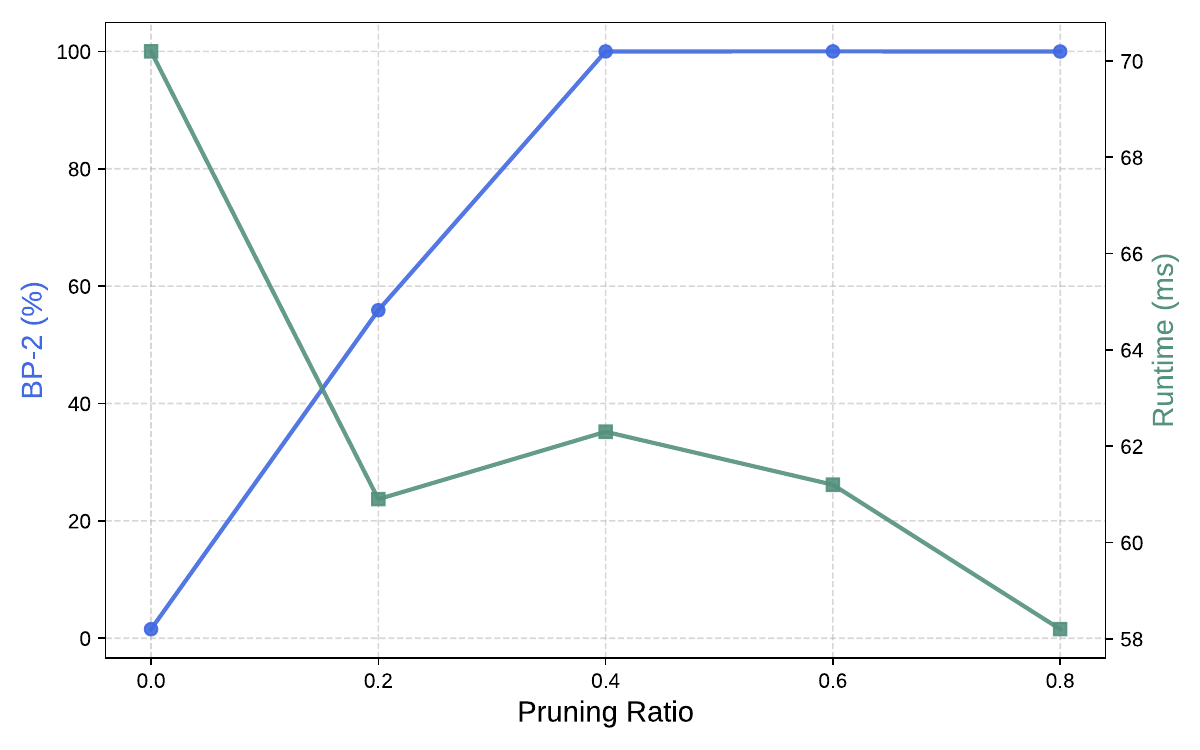}} 
    \vspace{-20pt}
    \caption{Study of applying structured pruning to cost filtering module, which is an alternative strategy to our introduced blockwise architecture search.} \label{fig:cost_volume_prune}
    \vspace{-15pt}
\end{figure}

\boldparagraphstart{Cost Filtering Pruning Study.}  As an alternative option to our introduced blockwise architecture search, we also experimented with directly applying structured pruning to the cost filtering module. As shown in  Fig.~\ref{fig:cost_volume_prune}, such strategy leads to marginal speed gain while substantially compromises the prediction accuracy. This degradation is likely attributable to the inherently small channel dimension (typically below 100) in cost volumes, which offers limited opportunities for effective pruning, in contrast to the significant redundancy in refinement module where structured pruning is more effective.

\section{Effects of Refinement Iterations}

\begin{figure}[h]
    \centering
    {\includegraphics[width=0.48\textwidth]{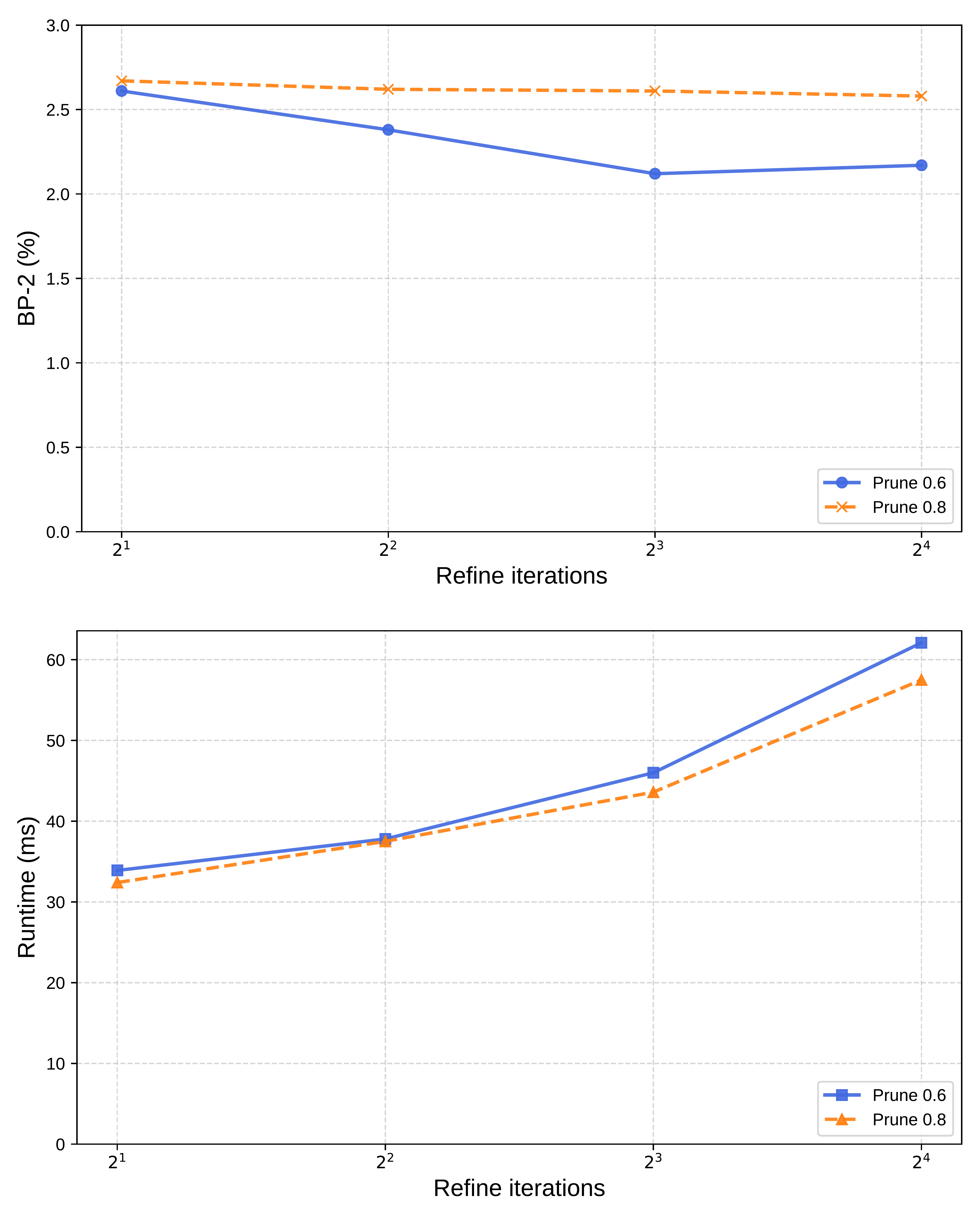}} 
    \vspace{-20pt}
    \caption{Effects of refinement iterations on accuracy (top) and runtime (bottom) under two different pruning ratios.} \label{fig:refine_iter_effect}
    \vspace{-15pt}
\end{figure}

Fig.~\ref{fig:refine_iter_effect} shows the effects of refinement iterations on accuracy and runtime of the complete model, where the results are evaluated on Middlebury-Q dataset. We compare two different versions of refinement module pruned with ratios of 0.6 and 0.8, while the remaining parts of the network (\eg feature extraction and cost filtering) are the same. As can be observed, under pruning ratio 0.8, the accuracy only obtains marginal improvements with increasing iterations steps, implying the aggressively pruned refinement module has less capacity to benefit from iterative refinement. Under pruning ratio 0.6, the accuracy saturates around 8 iterations. In terms of runtime, when the iteration steps are small, different pruning ratios lead to marginal differences, falling within the magnitude of milliseconds and introducing measurement noise. However, as the refinement iterations increase, higher pruning ratio yields more runtime benefit due to the cumulated effect. For our model evaluated in Table~\ref{tab:zero_shot} (main paper) pruning ratio is set to 0.6.

\section{More details on Model Efficiency}
Table~\ref{tab:more_runtime} shows more details of model efficiency analysis including runtime profiled on different NVIDIA GPUs. Our model obtains dramatic  reduction in parameters, MACs (Multiply-Accumulate Operations) and runtime across different hardware.

\input{table/more_runtime}

\section{Efficient GWC Volume Construction}
Group-wise correlation (GWC) volume has been widely used in prior works~\cite{wen2025foundationstereo,igev,igevpp,xu2022attention,shen2022pcw,bangunharcana2021correlate} to construct discriminative matching cost representations. Its original implementation~\cite{guo2019group} constructs the GWC volume by iterating over each disparity level in a Python for-loop: at each disparity $d$, the reference feature map is shifted relative to the target by $d$ pixels along the width axis, and a group-wise normalized dot product is computed between the aligned features, writing one disparity slice of the 5-D volume $\mathcal{V} \in \mathbb{R}^{B \times G \times D \times H \times W}$ per iteration. This loop incurs significant overhead from repeated GPU kernel launches (one per disparity level) and prevents the compiler from     
  fusing operations across disparity. Our optimized pytorch variant eliminates this explicit loop entirely by first left-padding the target feature map by $D{-}1$
   pixels and then unfold along the width dimension to extract all $D$ shifted views as a single strided tensor—a zero-copy operation that creates sliding windows over
  the padded data without materializing new memory. After a flip and permutation to align the disparity ordering, both the reference (broadcast-expanded via unsqueeze) and the unfolded target
  volume are reshaped into groups of $C/G$ channels, and correlated via a single fused element-wise multiply-and-sum.
  When combined with compilation of Pytorch or TensorRT, this formulation allows the compiler to trace and fuse the entire volume construction and correlation into a minimal number of GPU kernels, substantially
  reducing kernel launch overhead and improving memory access patterns compared to the original per-disparity-slice implementation. On image resolution of Middlebury-Q, we observe about $6\times$ runtime reduction, and $3\times$ memory usage reduction for constructing GWC volume. Our detailed implementation is available at: \url{https://github.com/NVlabs/Fast-FoundationStereo}

\section{Model Parameter and Memory Usage}
Table~\ref{tab:model_param} details parameter count comparison. Our model's peak memory usage is 0.63GB when running on Middlebury-Q. This fits into edge compute devices such as NVIDIA Jetson Orin or Thor, which are commonly used in real-time deployment for robotics and self-driving.
\input{table/param_compare}

\section{More Generalization Results}

Fig.~\ref{fig:zero_shot1} and \ref{fig:zero_shot2} demonstrate more qualitative results of zero-shot inference on out-of-domain stereo images featuring challenges such as textureless regions, translucent surfaces, specular highlights, diverse depth ranges, complex illuminations, varying viewing perspectives and indoor / outdoor scenarios.

\clearpage

\begin{figure*}[p]
    \centering
    {\includegraphics[width=0.99\textwidth]{fig/zero_shot1.pdf}} 
    \vspace{-5pt}
    \caption{Qualitative comparison among real-time methods. Results are obtained by zero-shot inference without training on target domain~\cite{wang2019flickr1024}. $^\dagger$Denotes training on the exact same datasets as ours (including our proposed pseudo-labels). Our pseudo-labeled internet data consistently enhances the generalization across different methods. However, our model demonstrates strongest robustness, validating  the effectiveness of both our model design and pseudo-labeling. (Zoom-in on a digital device for better visualization.)}\label{fig:zero_shot1}
    \vspace{-10pt}
\end{figure*}

\begin{figure*}[p]
    \centering
    {\includegraphics[width=0.99\textwidth]{fig/zero_shot2.pdf}} 
    \vspace{-5pt}
    \caption{Qualitative comparison among real-time methods. Results are obtained by zero-shot inference without training on target domain (images are from \cite{khazatsky2024droid} or captured in-the-wild). $^\dagger$Denotes training on the exact same datasets as ours (including our proposed pseudo-labels). Our pseudo-labeled internet data consistently enhances the generalization across different methods. However, our model demonstrates strongest robustness, validating  the effectiveness of both our model design and pseudo-labeling. (Zoom-in on a digital device for better visualization.)}\label{fig:zero_shot2}
    \vspace{-10pt}
\end{figure*}

\begin{figure*}[p]
    \centering
    {\includegraphics[width=0.99\textwidth]{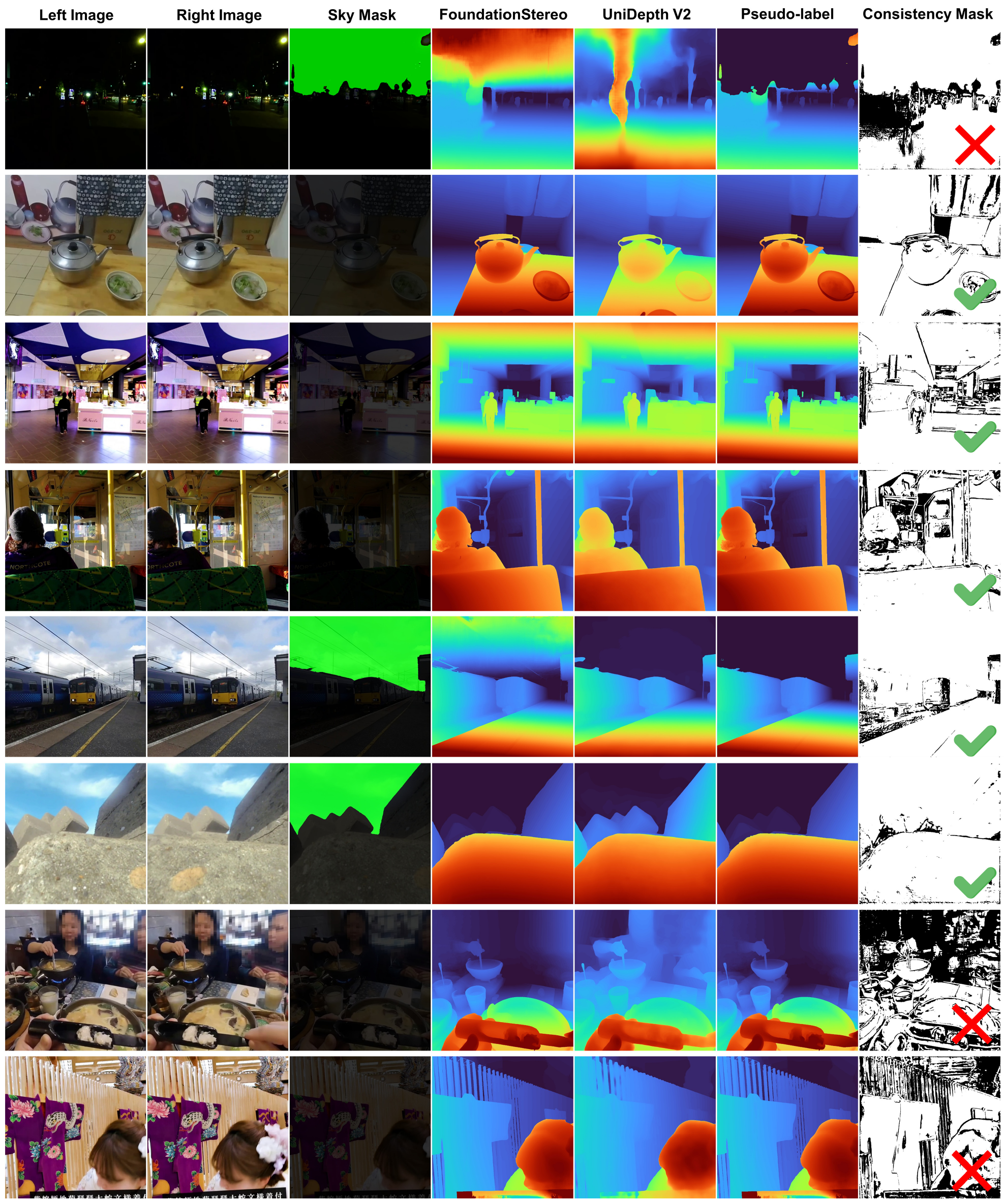}} 
    \vspace{-10pt}
    \caption{Visualizations of the intermediate results in our pseudo-labeling process. In the rightmost column, \includegraphics[height=1em]{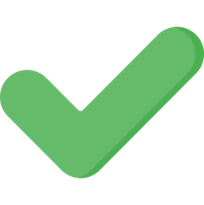}  or \includegraphics[height=1em]{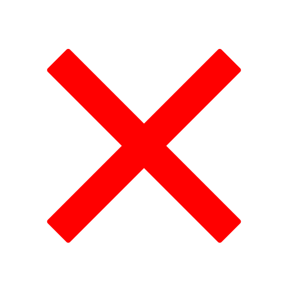} denotes whether samples are kept for training or not, based on the percentage of positive pixels in the consistency mask. Our data curation process can automatically discover failures on noisy internet data such as images containing subtitle (bottom), mosaic (2nd last row) and overly challenging samples that are unsuitable for training (top). The final pseudo-labels can also correct erroneous predictions from FoundationStereo on sky regions (5th row). (Zoom-in on a digital device for better visualization.)}\label{fig:stereo4d_samples}
    \vspace{-10pt}
\end{figure*}

\section{More Details on Pseudo-labeling}
Fig.~\ref{fig:stereo4d_samples} visualizes intermediate results in our pseudo-labeling process. Our data curation process can automatically discover failures on
noisy internet data such as images containing subtitle, mosaic and overly challenging samples that are unsuitable for training. 
The final pseudo-labels can also correct erroneous predictions from FoundationStereo on sky regions. Samples with positive pixels occupying more than $60\%$ (excluding sky regions) on the consistency mask are kept for training,  which results in 1.4M stereo pairs in total.

\section{Limitations}
While our method achieves strong generalization, it inevitably inherits certain limitations from its teacher FoundationStereo. Specifically, performance on translucent surfaces remains a challenge (Table~\ref{tab:booster} in main paper), which can be mitigated by incorporating training datasets enriched with relevant objects.

\section{Acknowledgement} 
We would like to thank Xutong Ren, Karsten Patzwaldt, Yonggan Fu, Saurav Muralidharan, Han Cai, Pavlo Molchanov, Yu Wang, Varun Praveen, Joseph Aribido and Jun Gao for their insightful early discussions for this project. We would also like to thank NVIDIA Isaac and TAO teams for their tremendous engineering support and valuable discussions.

%% file: table/more_runtime.tex
\begin{table}[h]
\centering
\def\mywidth{0.48\textwidth} 
\definecolor{green}{RGB}{0,200,0}
\resizebox{\mywidth}{!}{

\begin{tabular}{l *{5}{r}}
\toprule
\textbf{Method} & \multicolumn{3}{c}{\textbf{Runtime (ms)}} & \multirow{2}{*}{\textbf{\#Param (M)}} & \multirow{2}{*}{\textbf{MACs (G)}} \\
\cmidrule(lr){2-4}
& \textbf{3090} & \textbf{4090} & \textbf{A100} & & \\
\midrule
FoundationStereo~\cite{wen2025foundationstereo} & 496 & 295 & 308 & 374.5 & 5413.9 \\
Ours & 49 & 30 & 41 & 14.6 & 309.9 \\
\bottomrule
\end{tabular}

}
\vspace{-8pt}
\caption{Model efficiency analysis including runtime profiled on different NVIDIA GPUs. Ours corresponds to the model evaluated in Table~\ref{tab:zero_shot} (main paper).}\label{tab:more_runtime}
\vspace{-15pt}
\end{table}

%% file: table/param_compare.tex
\begin{table}[h]
\centering
\def\mywidth{0.45\textwidth} 
\definecolor{green}{RGB}{0,200,0}
\resizebox{\mywidth}{!}{

\begin{tabular}{lcccccc}
    \toprule
    Model & BANet-3D & BANet-2D & RT-IGEV & IINet & LightStereo-L & \textbf{Ours} \\
    \midrule
    Param (M) & 3.63 & 5.46 & 4.17 & 19.56 & 24.29 & \textbf{17.65} \\
    \bottomrule
\end{tabular}

}
\vspace{-8pt}
\caption{Model parameter comparison.}\label{tab:model_param}
\end{table}